\documentclass[sigconf,anonymous=false]{acmart}
\renewcommand\footnotetextcopyrightpermission[1]{}
\usepackage{booktabs}
\usepackage{xcolor}
\usepackage{dirtytalk}
\usepackage{graphicx}
\graphicspath{ {./images/} }
\usepackage{subcaption}
\usepackage{multirow}
\usepackage{tabularx}
\usepackage{array}
\usepackage{float}
\usepackage{multicol}
\usepackage[export]{adjustbox}
\usepackage{chngcntr}
\usepackage[title]{appendix}
\usepackage{booktabs}

\AtBeginDocument{%
  \providecommand\BibTeX{{%
    \normalfont B\kern-0.5em{\scshape i\kern-0.25em b}\kern-0.8em\TeX}}}

\setcopyright{acmcopyright}
\copyrightyear{2023}
\acmYear{2023}
\acmDOI{XXXXXXX.XXXXXXX}

\acmConference[VAT Workshop HRI'2023]{Make sure to enter the correct
  conference title from your rights confirmation email}{March 13, 2023}{Stockholm, Sweden}
\acmBooktitle{HRI '2023: Variable Autonomy for human-robot Teaming (VAT) workshop,
 March 13, 2023, Stockholm, Sweden} 

\begin{document}

\title{Robot Health Indicator: A Visual Cue to Improve Level of Autonomy Switching Systems}


\author{Aniketh Ramesh}
\email{Axr1050@student.bham.ac.uk}
\affiliation{%
  \institution{Extreme Robotics Laboratory, University of Birmingham}
  \streetaddress{Edgbaston}
  \city{Birmingham}
  \country{England}
  \postcode{B15 2TT}
}

\author{Madeleine Englund}
\email{Maen0191@student.umu.se}
\author{Andreas Theodorou}
\email{Andreas.Theodorou@umu.se}
\affiliation{%
  \institution{Umeå university}
  \streetaddress{MIT-huset}
  \city{Umeå}
  \country{Sweden}
  \postcode{901 87}
}

\author{Rustam Stolkin}
\email{R.Stolkin@bham.ac.uk}
\author{Manolis Chiou}
\email{M.Chiou@bham.ac.uk}
\affiliation{%
  \institution{Extreme Robotics Laboratory,\\ University of Birmingham}
  \streetaddress{Edgbaston}
  \city{Birmingham}
  \country{England}
  \postcode{B15 2TT}
}


\renewcommand{\shortauthors}{Ramesh, et al.}

\begin{abstract}
Using different Levels of Autonomy (LoA), a human operator can vary the extent of control they have over a robot's actions. LoAs enable operators to mitigate a robot's performance degradation or limitations in the its autonomous capabilities. However, LoA regulation and other tasks may often overload an operator's cognitive abilities. Inspired by video game user interfaces, we study if adding a \lq Robot Health Bar\rq \  to the robot control UI can reduce the cognitive demand and perceptual effort required for LoA regulation while promoting trust and transparency. This Health Bar uses the robot vitals and robot health framework to quantify and present runtime performance degradation in robots. Results from our pilot study indicate that when using a health bar, operators used to manual control more to minimise the risk of robot failure during high performance degradation. It also gave us insights and lessons to inform subsequent experiments on human-robot teaming.
\end{abstract}

\maketitle

\section{Introduction and Related Work}

During mobile robot navigation tasks, factors like uneven terrain, obstacles, sensor noise etc. can degrade a robot's performance. If left unattended, performance degradation can cause robots to fail or perform sub-optimally \cite{ramesh2022robot}. Timely operator intervention or triggering of recovery behaviours can mitigate this. Using Levels of Autonomy (LoA) \cite{sheridan1978human}, the extent of control that a human operator has over the robot's actions can be varied. For example during high performance degradation, an autonomous robot can be switched to manual control by an experienced human operator. The LoA can be switched back once the robot is capable of functioning autonomously again.

In Human Initiative (HI) LoA switching systems, the robot operator is in charge of LoA regulation during the task. Compared to fully autonomous robots, manually controlled robots, and other implementations of variable autonomy systems, HI-LoA shows better task performance during remote navigation tasks in unknown environments \cite{chiou2016experimental}. However, the additional perceptual and mental effort required to monitor robot operation data and determine if LoA switching is needed, imposes comparatively higher levels of cognitive workload on the operator. While overloading an operator's cognitive abilities for prolonged periods can reduce task performance \cite{prewett2010workload} due to stress, fatigue and varying levels of trust in the system \cite{mizuno2011fatigue, agrawal2018trust}, low workload can cause out-of-loop performance problems like complacency and over-trusting the system \cite{fendley2017cognitiveload, hussein2018cogload}.
Effective HI-LoA system design remains an open problem in existing literature \cite{chiou2016experimental}. A well-designed system should keep operator cognitive workload within a \lq sweet spot \rq \ \cite{ashcraft2019moderating} (i.e., acceptable levels). LoA switches by operators should be due to a clear understanding of the robot's capabilities and limitations, not due to trust issues. Transparent design, i.e., where the operator can accurately interpret the robot's capabilities, goals, and its progress, have reported effective calibration of trust \cite{vitale2018more}, even with increasing levels of automation \cite{chen2014human}. Systems where information about the robot is presented over the graphic/visual modalities are trusted more than the audio and textual modalities \cite{Sanders2014modalityandtransparency}. Independent of systems design considerations, significant differences still exist in people's perceived reliability of variable autonomy systems. However, they can be minimised through standardised training \cite{chen2014human}. 

Video games serve as inspiration to understand how real-time information can be presented in a transparent manner, while keeping the operator cognitive workload in the sweet spot. Video game players report that they are receptive to information presented through visual cues as long as they are clear and consistent \cite{llanos2011intergratedUI}. Game visual cues are usually colour coded UI elements overlayed on the interface during gameplay, e.g. the \lq Health Bar\rq. In popular online games\footnote{See https://overwatch.blizzard.com/en-us/}, remotely situated players use these cues to get their teammates attention. Similarly, robot operators can use colour coded visual cues to acquire quick situational awareness \cite{steinfeld2004interface} while operating a robot with minimum perceptual effort. For robot interfaces, Murphy et al. \cite{murphy2019user} recommend that visual cues may be added to the robot control UI, without blocking useful information.

\begin{figure*}[t]
  \includegraphics[width=\textwidth]{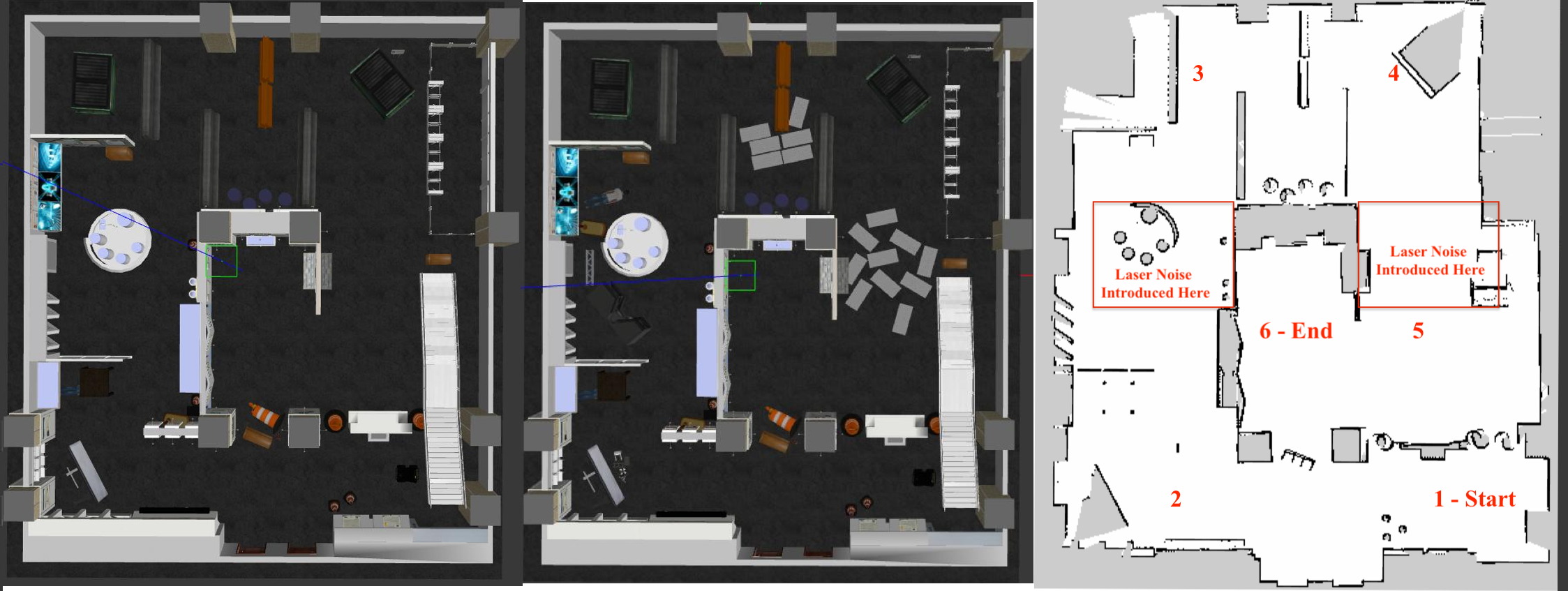}
  \caption{Condition A arena (L to R): 1) Empty, 2) With obstacles and uneven terrain, 3) 2D map with waypoints marked and locations where laser noise was introduced.}
  \label{fig:arena}
\end{figure*}

The existing literature shows that effective trust calibration, system transparency and standardised training are crucial for a well designed HI-LoA switching system. Information to assist the operator with making decisions about LoA regulation can be presented through properly designed visual cues. Therefore, in this study, we present our pilot experiments to explore how visual cues about a robot's performance degradation can help LoA regulation decisions during high cognitive workload situations. Based on the Robot Vitals and Robot Health framework \cite{ramesh2022robot} to quantify performance degradation, we design a colour coded \lq Robot Health Bar \rq \ UI element. We hypothesise that using the Health Bar will reduce the perceptual effort required by operators to make LoA switching decisions, improve task performance, and result in a lower overall cognitive workload. We also took feedback from the participants about their overall trust in the system, its transparency and recommendation for improving the Health Bar design.

\section{System Design}

The experiment consisted of two tasks, similar to the experiments setup by Chiou et al. \cite{chiou2016experimental,chiou2021mixed}. The primary task was a mobile robot navigation task using the Clearpath Husky Robot, simulated on Gazebo. As shown in Figure \ref{fig:arena}, these arenas were designed to mimic urban search and rescue scenarios, and were populated with performance degrading factors commonly found in them like - obstacles, uneven terrain and laser noise. A 2D laser scan of an empty map (Figure \ref{fig:arena} - Right) was generated before the performance degrading factors were added. The difference between the map used for robot navigation and the actual arena affects robot navigation planning, thereby adding another performance degrading factor. Two different arenas were created to compare operator performance with (condition A) and without the Health Bar (condition B). 

\begin{figure}[H]
  \centering
    \includegraphics[width=\linewidth]{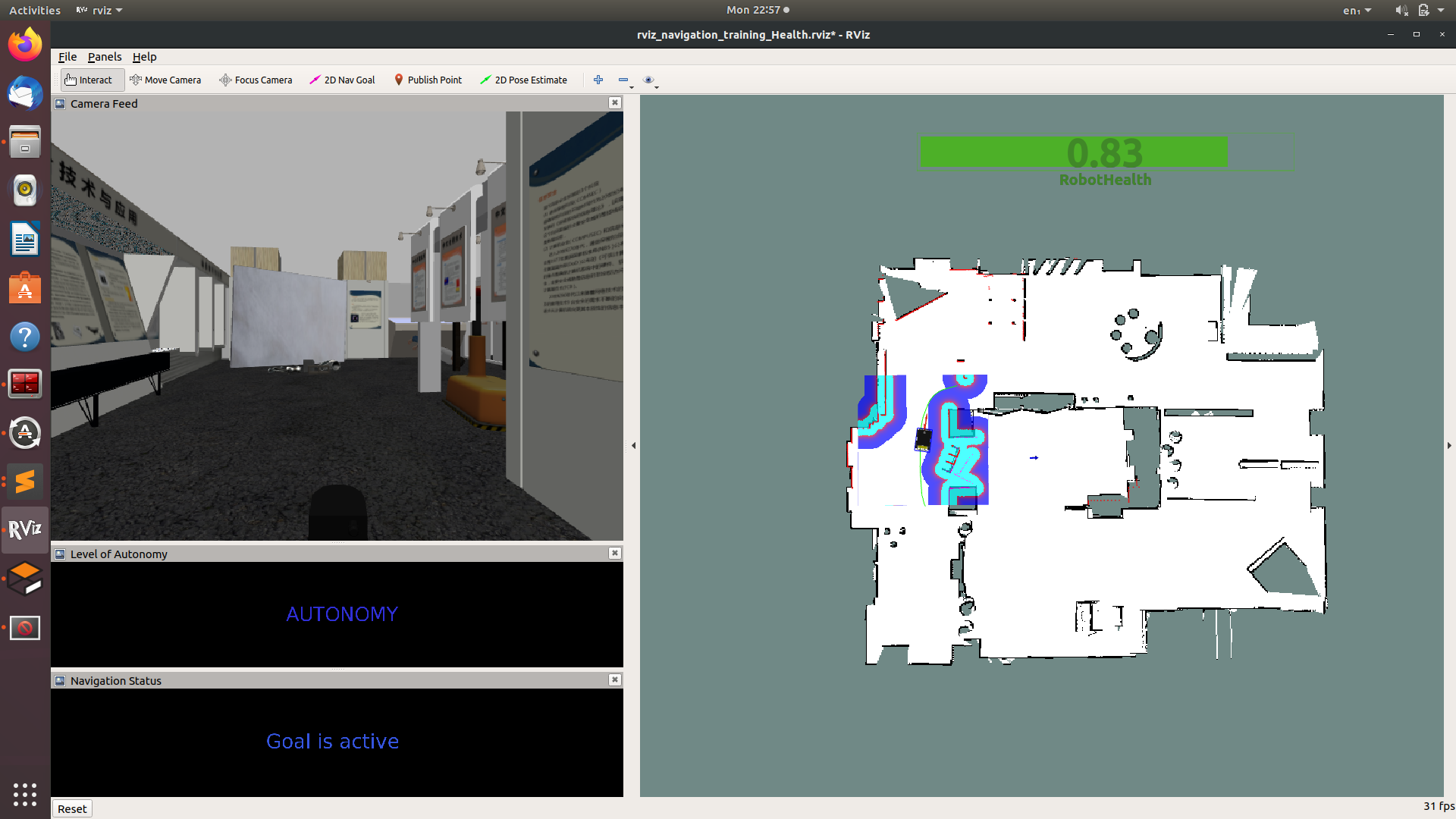}
    \caption{Condition A - Interface with Robot Health Bar. The green colour indicates the robot is 'healthy'}
    \label{fig:UI1}
\end{figure}

RViz was used to visualise the robot map, sensor data, and give commands to the robot. The standard RViz interface in Figure \ref{fig:UI2} was used for Condition B, and the interface with a \lq Robot Health Bar \rq \ shown in Figure \ref{fig:UI1} was used for Condition A. The Husky robot used two LoAs - 1) Waypoint-based navigation and 2) Manual Control by an operator using a Joystick. The Robot Health at each instant was calculated using the Robot Vitals and Robot Health framework \cite{ramesh2022robot}, and the Health Bar was created and displayed using the JSK Visualisations\footnote{http://wiki.ros.org/jsk\_visualization/} ROS Package. Here, Robot Health is defined as \lq an overall scalar estimate of a robot’s ability to carry out its tasks without its capabilities being impaired by any performance degrading factors \rq. Therefore, an operator can monitor the health to detect situations where a robot is likely to fail, and trigger LoA switches to assist the robot. To improve readability, the Robot Health was standardised to the range $[0,1]$. The Red-Amber-Green colour coding convention was used for the Health Bar as shown in Figure \ref{fig:UI1}. Based on preliminary experiments, health above $0.7$ was decided as \lq Healthy \rq, and the colour of the Health Bar was green. Values between $0.5-0.7$ were coloured in amber, and values below $0.5$ were coloured red. Instead of sharp colour changes, the Health Bar gradually went from green to amber to red. This was done to minimise rapid colour changes due to small fluctuations in health near the threshold values.

The secondary task was a 3D object rotation task \cite{Ganis2016}, used to induce additional cognitive workload for the operator. Here participants were successively presented with two 3D objects. They had to determine whether the objects were the same or different. Instead of having a fixed time limit like in Chiou et al.\cite{chiou2021mixed}, here the participants did this task throughout the runtime.  Both the primary and secondary tasks were displayed on adjacent screens. Participants were explicitly instructed \lq to prioritise the robot navigation task and minimise the likelihood of robot failure \rq. They were also told to simultaneously do the secondary task to the best of their ability, and that they were not being evaluated on their performance in it.

\begin{figure}[H]
    \includegraphics[width=\linewidth]{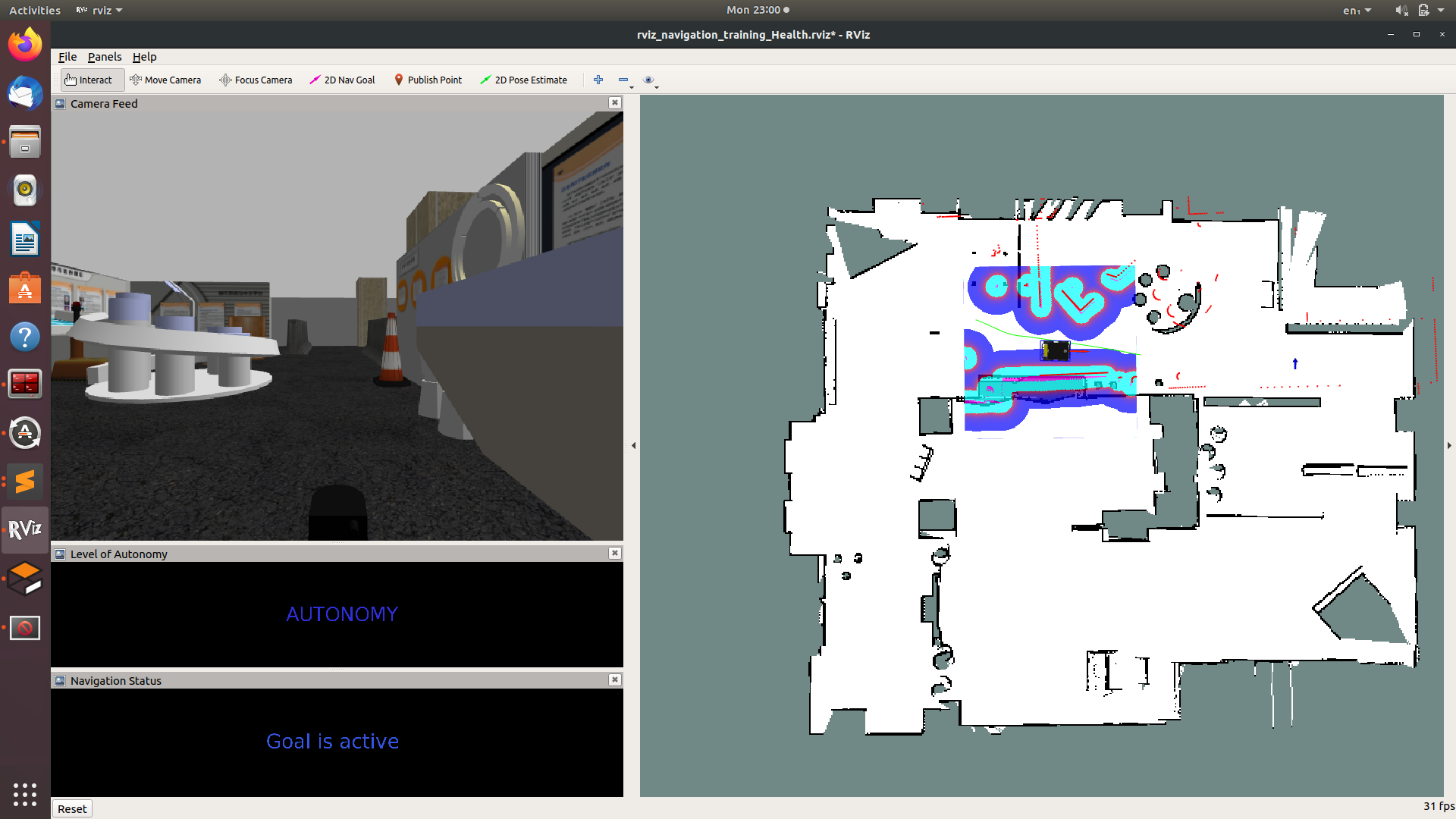}
    \caption{Condition B - Interface without Robot Health Bar}
  \label{fig:UI2}
\end{figure}

\section{Experiment Methodology}
Eight test subjects participated in the experiment and performed both conditions (i.e. within subjects design). The order of the conditions was counterbalanced to minimise learning and fatigue effects. First, participants had to fill out background information (see appendix \ref{background}). Then they were introduced to basic robot navigation in a training arena similar to the one used in Chiou et al.\cite{chiou2021mixed}. All participants trained on this arena till they were able to demonstrate a minimum proficiency in robot navigation. This ensured that confounding factors due to a variation in skill levels were minimised. Next, they were shown the 2 different LoA, how to switch between them and were given some time to practice LoA switching on the training arena. 

Participants were then introduced to the interface with the Health Bar and asked open-ended questions without prior explanation (see appendix \ref{open-ended qs}). After answering the questions, they were given the following instructions regarding the Health Bar: “The Robot Health Bar indicates how much a robot’s performance is degraded by environmental factors. These environmental factors can be anything ranging from bad terrain to laser noise. During low health, the Health Bar will become redder. During high health, the Health Bar will become greener. The lower the robot's health, the more likely it will fail. You may use the Health Bar to help you determine when the robot requires a LoA switching”. The effect of performance degrading factors on the Health Bar was demonstrated by introducing obstacles and laser noise in the training arena. Participants were also given time to familiarise themselves with the interface before starting the navigation task. Next participants were familiarised only with the secondary task, and allowed multiple practice runs till they felt comfortable with it. Their baseline on the secondary task alone was measured before the experiment.


Before each condition, participants were shown the start and finish points in the navigation task and told that the next waypoint would be assigned automatically on reaching the current one. Then the participants carried out each experimental condition followed by a NASA-TLX form to evaluate the perceived cognitive workload during the task. Lastly, participants completed open-ended questions and a transparency/trust questionnaire after the experimental trials and the NASA-TLX forms (see \ref{open-ended qs} and \ref{transparency qs}).


\begin{table}[]
\centering
\caption{Summary of Statistical Analysis}
\resizebox{\columnwidth}{!}{%
\renewcommand{\arraystretch}{1.75}
\large
\begin{tabular}{@{}ccccl@{}}
\toprule
\textbf{Primary Task}                                        & \textbf{Condition A - With Health Bar} & \textbf{Condition B -Without Health Bar} & \textbf{Pairwise T Test, Two Tailed} &  \\ \midrule
Completion Time                                              & M 174.38,  SD = 30.91                  & M 175.25,  SD = 32.35                    & T-Score = 2.36, P Value = 0.96       &  \\
Percentage of the Task Robot was Manually Controlled            & M 49.13,  SD = 19.97                   & M 39.77,  SD = 17.77                     & T-Score = 2.36, P Value = 0.08       &  \\
Percentage of the Task Robot was Unhealthy                   & M 61.91,  SD = 21.80                   & M 78.03,  SD = 21.36                     & T-Score = 2.36, P Value = 0.17       &  \\
Percentage of the Task Robot was Manually Controlled when Unhealthy & M 64.59,  SD = 21.16                   & M 45.17,  SD = 18.67                     & T-Score = 2.36, P Value = 0.02*       &  \\ \midrule
\textbf{Secondary Task}                                      & \textbf{Condition A - With Health Bar} & \textbf{Condition B -Without Health Bar} & \textbf{Pairwise T Test, Two Tailed} &  \\ \midrule
Average Response Time                                        & M 10,616.18,  SD = 3,673.76            & M 9,053.49,  SD = 3,679.02               & T-Score = 2.36, P Value = 0.22       &  \\
Total Answered                                               & M 20.63,  SD = 5.80                    & M 23.63,  SD = 8.35                      & T-Score = 2.36, P Value = 0.19       &  \\
Accuracy \%                                                  & M 90.13,  SD = 7.00                    & M 94.59,  SD = 4.98                      & T-Score = 2.36, P Value = 0.02*       &  \\ \midrule
\textbf{NASA-TLX}                                            & \textbf{Condition A - With Health Bar} & \textbf{Condition B -Without Health Bar} & \textbf{Pairwise T Test, Two Tailed} &  \\ \midrule
Mental Demand                                                & M 80.63,  SD = 16.78                   & M 80.00,  SD = 17.73                     & T-Score = 2.36, P Value = 0.89       &  \\
Physical Demand                                              & M 28.75,  SD = 13.30                   & M 36.88,  SD = 25.35                     & T-Score = 2.36, P Value = 0.19       &  \\
Temporal Demand                                              & M 66.88,  SD = 26.98                   & M 70.00,  SD = 17.11                     & T-Score = 2.36, P Value = 0.60       &  \\
Performance                                                  & M 43.75,  SD = 29.73                   & M 43.13,  SD = 24.63                     & T-Score = 2.36, P Value = 0.94       &  \\
Effort                                                       & M 58.75,  SD = 30.21                   & M 59.38,  SD = 27.83                     & T-Score = 2.36, P Value = 0.91       &  \\
Frustration                                                  & M 49.38,  SD = 28.21                   & M 53.13,  SD = 27.51                     & T-Score = 2.36, P Value = 0.36       &  \\ \bottomrule
\multicolumn{3}{p{45.915em}}{*p<0.05}
\end{tabular}%
}
\label{tab:stats1}
\end{table}

\section{Results}

Robot health values under $0.7$ were classified as \lq unhealthy \rq. This criterion was heuristically determined based on previous studies on the robot vitals and robot health framework \cite{ramesh2022robot}. Results from the experiments conducted on 8 participants were computed and tested for statistical differences using two-tailed pairwise T-Tests. These results are summarised in table \ref{tab:stats1}.

The total percentage of runtime that the robot was autonomous, manually controlled and \lq unhealthy \rq \ was calculated for each experimental condition. The percentage of run time that the robot health was \lq unhealthy \rq \ showed no statistical differences between the two conditions. Similarly, no significant differences were observed in the percentage of run time the robot was manually controlled. However, the percentage of run time that the robot was manually controlled when it was unhealthy, showed a significant ($p <0.05$, $t = -2.36$) difference between both conditions. That is, operators manually controlled unhealthy robots for longer when the health of the robot was displayed (i.e., Condition A). As a result, the robot was \lq unhealthy \rq \ for an average of $61.91\%$ of the runtime in condition A, but $78.03\%$ of the runtime in condition B.

The change in perceptual effort required to carry out LoA switching tasks was measured using the operator's accuracy on the secondary task. When the health bar was not displayed (Condition B), operators showed significantly higher levels of accuracy ($p <0.05$, $t = 2.36$) on the secondary task. This indicates the perceptual effort required to use the interface with the robot health bar, is higher. However, the NASA-TLX scores showed no significant differences in the overall cognitive workload imposed by both conditions.



Open-ended questions asked before the experiment showed that some participants were slightly confused about what the Robot Health Bar was. One participant said - "The health bar looks like a timer, because it has a number. I thought it meant seconds". Others successfully grasped the idea behind the UI element. Another participant thought "When the number goes down the robot dies", indicating a game-like perception of the Health Bar's function. However, all participants understood it was there to assist or alert an operator performance degradation.


Most participants preferred the interface of condition A (with the \lq Robot Health bar \rq), finding it intuitive for regulating LoA - "When the health was low, it was better to manually control the robot". One participant said - "Without the health bar I had to actively use my brain to detect when the robot needed help". Three participants, however, felt that the colours were sometimes inaccurate and did not match the robot's state. One of them pointed out that while red and green clearly helped indicate which LoA was better, amber was confusing because they did not know what to do. Although not significant, operators that had experience with operating robots (according to the background information) seemed to prefer more transparency in the UI, while the rest preferred less transparency. Novice operators (as reported by the background information), instead, stated that they responded to the colours of the UI to a greater extent than the participants with more experience in operating robots.


Pearson's correlation showed a significant negative trend between how often the participants operated remote-controlled vehicles and how easy it was to know when to change LoA for alternative 1 (Figure \ref{fig:low}) in the questionnaire ($p <0.01$). Significant positive trends were also observed for how often they used AI for work and how easy it was to know when to change LoA for alternative 3 in the questionnaire (Figure \ref{fig:high}) ($p <0.05$), and how easy it was to understand why to change LoA for alternative 2 (Figure \ref{fig:med}) ($p <0.01$) and alternative 3 in the questionnaire(Figure \ref{fig:high}) ($p <0.01$) (see Table \ref{tab:correlation}). 

In terms of determining when to change LoA for the alternatives in the questionnaire, four participants rated either the highest or both the lowest and highest levels of transparency equally highly on the Likert scale (Figure \ref{fig:when}). Conversely, the remaining four participants rated the lowest level of transparency highest in comparison to the other two levels. Overall, the medium level of transparency was rated the lowest. As for why to change LoA, six participants either increased or maintained their rating for each increase in the level of transparency, resulting in the highest level of transparency being rated the highest (Figure \ref{fig:why}). However, two participants decreased their rating on the Likert scale for each increase in the level of transparency.

\section{Discussion and Insights}

Our experimental results illustrate how informing an operator about a robot's performance degradation through visual cues can affect the operator's driving and LoA regulation style. When shown the health bar, operators triggered LoA switches to mitigate situations where the robot was \lq unhealthy \rq. This style of HI-LoA regulation reduces the aggregate risk of robot failure. The interface with the health bar did not impose a significant additional cognitive workload on the operator during the experiment, and did not significantly change the task completion time. Contrary to our hypothesis, LoA regulation using the Health Bar required significantly ($p<0.05$) higher perceptual effort. In condition A, Participants carried out fewer rotations and gave less accurate answers. One likely explanation for this is that adding a Health Bar increased the number of points on the UI the operator had to focus on, which increased the perceptual effort for the primary task.

Table \ref{tab:correlation} shows that novice robot prefer significantly less transparency than experienced robot operators. Participant feedback revealed that while novice operators responded more to the colours of the health bar, experienced operators preferred having more information displayed to inform their LoA switching decisions. This indicates that participants value transparency only when they understand the necessity of transparency. People with different levels of experience have varied mental models about the Robot, leading them to perceive the system and the Health Bar differently. This makes it difficult to standardise explanations. Depending on their experience, each participant may want to clarify different aspects of the Human-Robot System, leading to differences in the way they are primed for the experiment. Therefore, rigorous training coupled with detailed explanations about the different components of the system are required to minimise the differences in perception. Finally, the major limitation of this study is the small sample size. Therefore all the results and insights presented in this study can require validation with a larger set of participants.

\subsubsection*{Participant Recommendations to improve the Robot Health Bar and User Interface Design:}
\begin{itemize}
\item Make the Health Bar more Salient, so that it attract attention when necessary
\item Do not include a drop down-menu in the Health Bar as shown in the Questionnaire
\item Use a percentage to the health value instead of a value between $[0,1]$
\item  To attract attention to 'low health' use multi-modal awareness cues like sound alerts or UI elements e.g. By Shaking or Blinking the health bar after a threshold value.
\end{itemize}





\section{Conclusion and Future Work}

In this study, we explored if visual cues about a robot's performance degradation can reduce the perceptual effort required to make LoA switching decisions for remote mobile robot navigation tasks. Inspired by video games, we designed a 'Robot Health Bar'. This health bar displayed the total runtime performance degradation the robot is facing. A total of 8 participants carried out a mobile robot navigation task with and without the health bar UI element under high cognitive workload, and their performance was measured. Adding a Robot Health Bar to the robot control UI significantly changed how the operator makes Level of Autonomy Switching Decisions. When the Health Bar was displayed, operators were more attentive and took control of the robot more to minimise the risk of the robots failing. Visual cues that indicate a robot's health can serve as an effective way of ensuring safer control of robots, especially in extreme environments where robot missions have high levels of risk and environmental adversities.  In the future, we aim to explore how the insights gained from this study can scale to multi-robot systems.



\bibliographystyle{unsrtnat}
\bibliography{sample-base}

\appendix

\section{Open-ended questions and questionnaires}
\subsection{Background information questionnaire}\label{background} 
The background questionnaire consisted of the following questions:
\begin{enumerate}
\item[\textbf{Q1:}] How often do you operate or use to operate remote controlled vehicles (e.g. Robots, Drones, Heavy machinery)
\item[\textbf{Q2:}] How often do you play or used to play video games involving driving, flight simulation and third person shooters, RPG and sports?
\item[\textbf{Q3:}] Do you use AI (e.g. Autonomous Robots, Machine Learning Algorithms, AI Tools ) for work?
\item[\textbf{Q4:}] Do you use AI (e.g. Personal Assistant) in your personal life?
\end{enumerate}
The possible answers ranged between 1-5; i.e., least to most often.

\subsection{Open-ended questions}\label{open-ended qs} 
Questions asked before participants had performed the experiment under both conditions are listed below:
\begin{enumerate}
\item[\textbf{Q1:}] What are you thinking as you look at this?
\item[\textbf{Q2:}] What is your first impression of this UI element?
\item[\textbf{Q3:}] What do you think this UI element does or will do?
\end{enumerate} 

All questions asked after participants had performed the experiment under both conditions are listed below:
\begin{enumerate}
\item[\textbf{Q1:}] Was anything surprising or did not perform as expected in either of the interfaces?
\item[\textbf{Q2:}] Was the interface without the \lq Robot Health Bar \rq \ easy to understand?
\item[\textbf{Q3:}] Was the interface with the health bar easy to understand?
\item[\textbf{Q4:}] What did you think about the colours used in the health bar when the health changed?
\item[\textbf{Q5:}] Was the LoA switching behaviour of the interface without the \lq Robot Health Bar \rq \ transparent?
\item[\textbf{Q6:}] Was the LoA switching behaviour of the interface with the \lq Robot Health Bar \rq \ transparent? \\
\item[\textbf{Q7:}] Is there anything else that you can think of, specific to the UI and UX to improve the use of LoA switching Robots?
\end{enumerate}

\subsection{Transparency/trust questionnaire}\label{transparency qs} 

The transparency/trust questionnaire consisted of images of three different alternatives of a health bar ranging from low to high transparency and a 7-point Likert scale for each image (Figure \ref{fig:low to high}). The questions were the following pair, for each alternative:
\begin{enumerate}
\item[\textbf{Q1:}] It is easy to understand WHEN to change Level of Autonomy (LoA) based ONLY on alternative 1?
\item[\textbf{Q2:}] It is easy to understand WHY to change or not change Level of Autonomy (LoA) based ONLY on alternative 1? 
\end{enumerate}

\newpage
\onecolumn
\section{results}
\setcounter{figure}{0}
\setcounter{table}{0}
\counterwithin{table}{section}
\counterwithin{figure}{section}

\begin{table}[h!]
    \centering
    \caption{Correlation Between Background and Online Questionnaire}
    \begin{tabular}{m{3cm} m{1.5cm} m{3cm} m{3cm} m{2.5cm} m{2cm}}
    \hline
    & 
    & 
    How often do you operate or use to operate remote-controlled vehicles (e.g. Robots, Drones, Heavy machinery) 
    & How often do you play or used to play video games involving driving, flight simulation and third person shooters, RPG and sports? 
    & Do you use AI (e.g. Autonomous Robots, Machine Learning Algorithms, AI Tools) for work? 
    & Do you use AI (e.g. Personal Assistant) in your personal life? \\
    \hline
         {It is easy to understand when to change Level of Autonomy (LoA) based ONLY on alternative 1?} & Pearson's r
         p-value & -0.874
         
         0.005** 
         & 0.045
         
         0.916 
         & -0.338
         
         0.413
         & -0.336
         
         0.416\\
         & & & & & \\
         
         {It is easy to understand when to change Level of Autonomy (LoA) based ONLY on alternative 3?} & Pearson's r
         p-value & 0.455 

         0.257
         & -0.277 
         
         0.507
         & 0.802 
         
         0.017*
         & 0.120 
         
         0.778\\
         & & & & & \\
         {It is easy to understand WHY to change or not change Level of Autonomy (LoA) based ONLY on alternative 2?} & Pearson's r 
         p-value & 0.363 
         
         0.376
         & 0.051 
         
         0.905
         & 0.895 
         
         0.003**
         & 0.191 
         
         0.651\\
         & & & & & \\
         {It is easy to understand WHY to change or not change Level of Autonomy (LoA) based ONLY on alternative 3?} & Pearson's r 
         p-value & 0.433 
         
         0.284
         & -0.266 
         
         0.524
         & 0.858

         0.006** & 0.000
         
         1.000\\
    \hline
    \multicolumn{6}{l}{*p \textless 0.05, **p \textless 0.01}\\
    \multicolumn{6}{l}{\textit{Note.} Only a few selected results from the Pearson's correlation are displayed}\\
    \end{tabular}
    \label{tab:correlation}
\end{table}

\begin{figure*}[h!]
    \setcounter{figure}{1}
    \subfloat[Results for the questions regarding when to change LoA]{\label{fig:when}
    \includegraphics[width=8cm, height=5cm]{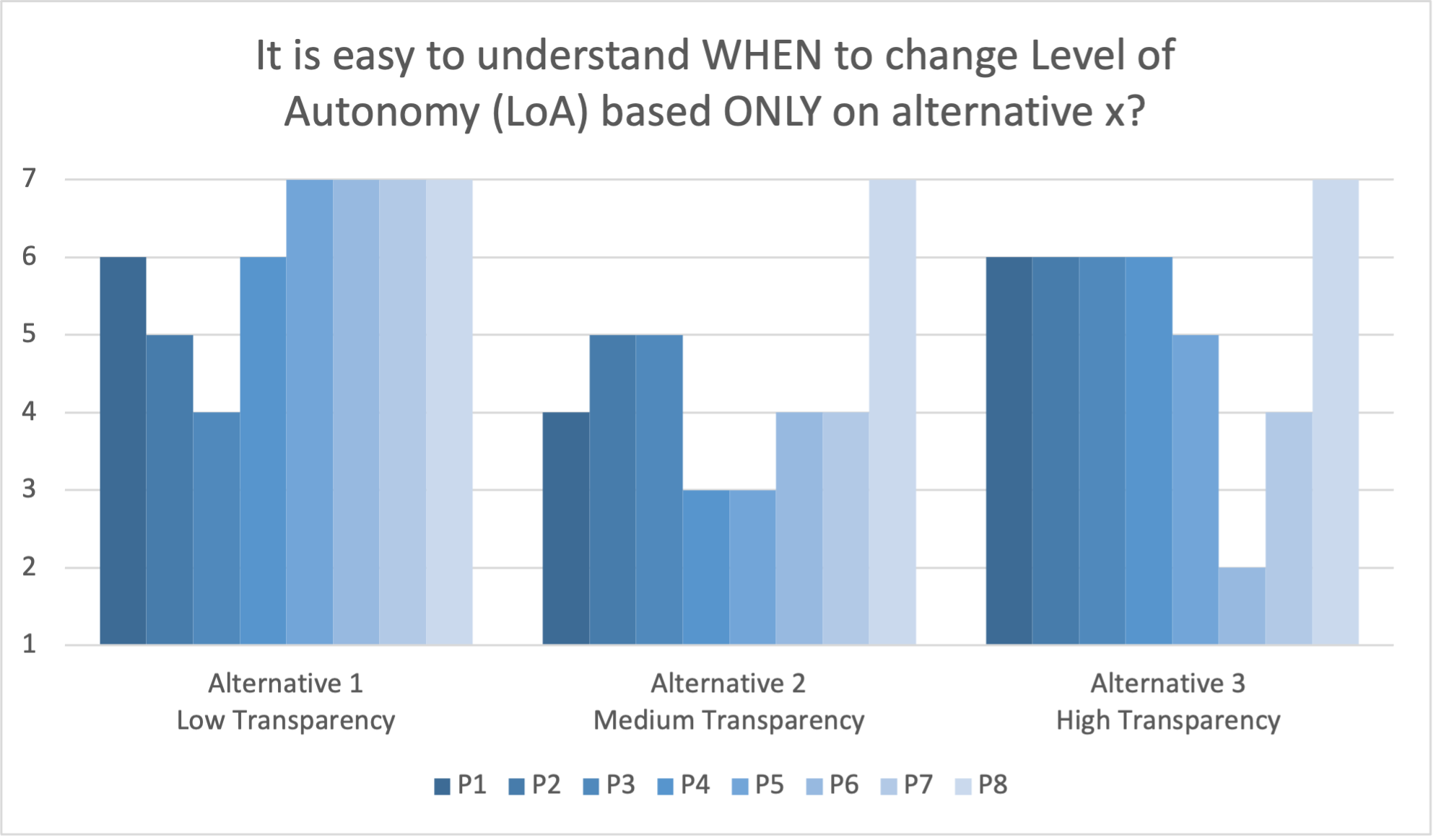} 
   }
    \subfloat[Results for the questions regarding why to change LoA]{\label{fig:why}
    \includegraphics[width=8cm, height=5cm]{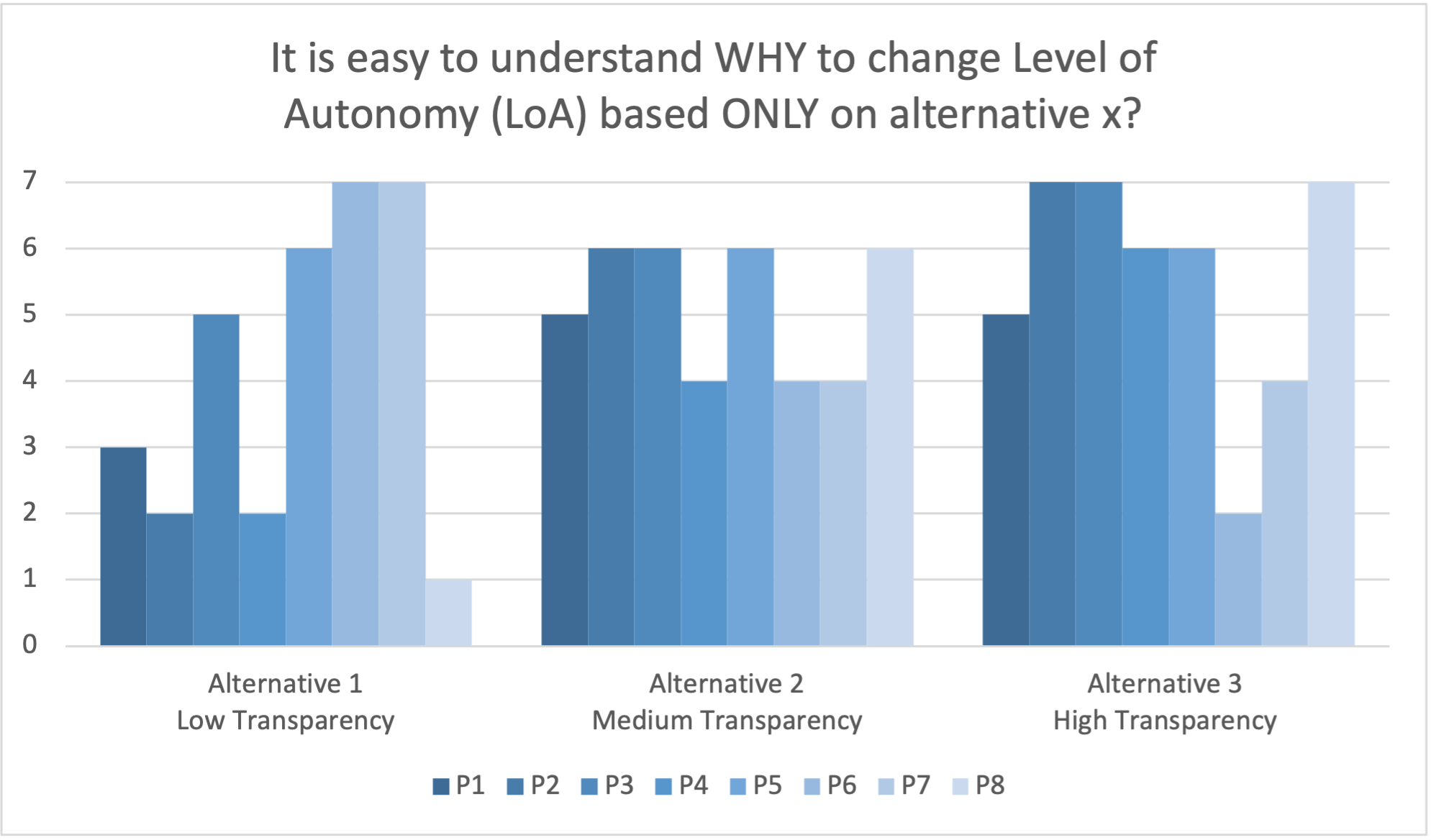} 
    }
\caption{Results from questionnaire}
\label{fig:results from questionnaire}
\end{figure*}

\begin{figure}[h!]
    \subfloat[Alternative 1 - \protect\\ Low transparency]{\label{fig:low}
    \includegraphics[width=6cm, height=3cm]{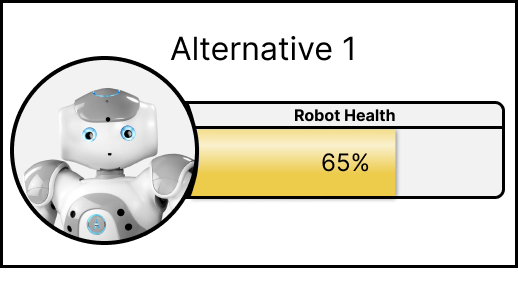} 
   }
    \subfloat[Alternative 2 - Medium transparency]{\label{fig:med}
    \includegraphics[width=6cm, height=6.4cm]{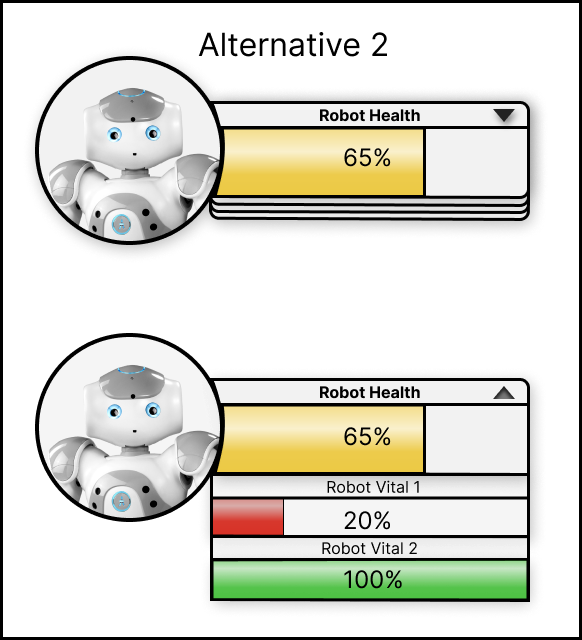} 
    }

    \subfloat[Alternative 3 - High transparency]{\label{fig:high}
    \includegraphics[width=12cm, height=6.4cm]{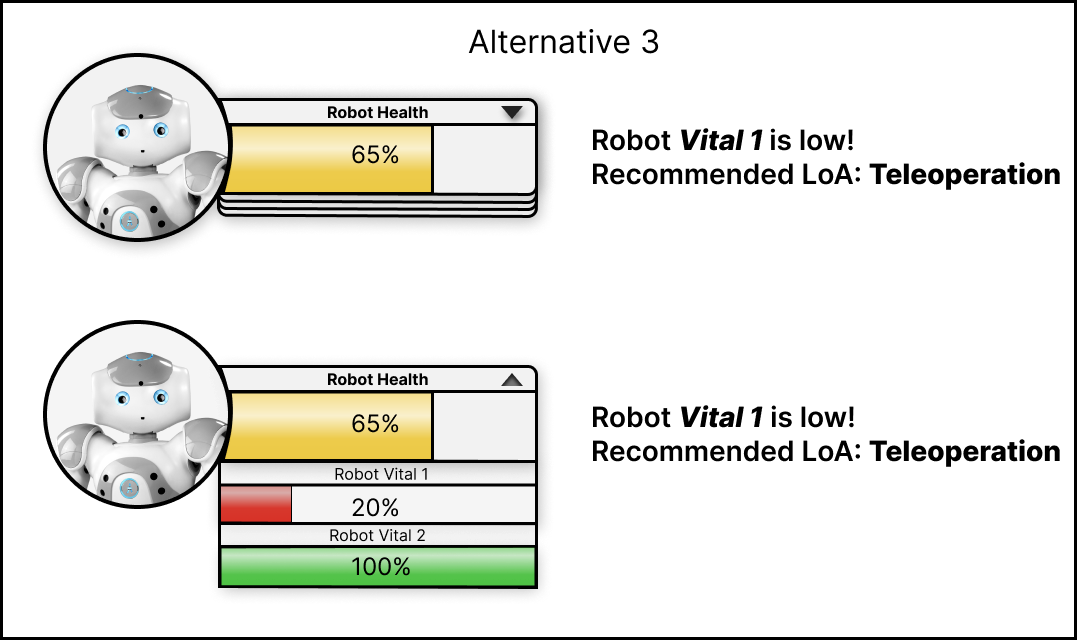} 
    }
\caption{Alternatives ranging from low to high transparency}
\label{fig:low to high}
\end{figure}

\end{document}